\documentclass{article}

\usepackage[english]{babel}

\usepackage[letterpaper,top=2cm,bottom=2cm,left=3cm,right=3cm,marginparwidth=1.75cm]{geometry}

\usepackage{amsmath}
\usepackage{graphicx}
\usepackage[colorlinks=true, allcolors=blue]{hyperref}

\title{Bridging the Gap: Deciphering Tabular Data Using Large Language Model}
\author{
  Hengyuan Zhang \and
  Peng Chang\thanks{Corresponding author. PAII Inc., Palo Alto, CA 94306, USA. Email: pengchang@gmail.com} \and
  Zongcheng Ji
}

\begin{document}
\maketitle

\begin{abstract}
In the realm of natural language processing, the understanding of tabular data has perpetually stood as a focal point of scholarly inquiry. The emergence of expansive language models, exemplified by the likes of ChatGPT, has ushered in a wave of endeavors wherein researchers aim to harness these models for tasks related to table-based question answering. Central to our investigative pursuits is the elucidation of methodologies that amplify the aptitude of such large language models in discerning both the structural intricacies and inherent content of tables, ultimately facilitating their capacity to provide informed responses to pertinent queries. To this end, we have architected a distinctive module dedicated to the serialization of tables for seamless integration with expansive language models. Additionally, we've instituted a corrective mechanism within the model to rectify potential inaccuracies. Experimental results indicate that, although our proposed method trails the SOTA by approximately 11.7% in overall metrics, it surpasses the SOTA by about 1.2% in tests on specific datasets. This research marks the first application of large language models to table-based question answering tasks, enhancing the model's comprehension of both table structures and content.
\end{abstract}

\section{Introduction}

Tabular data forms the cornerstone of numerous sectors, ranging from healthcare and finance to marketing and machine learning. Its pervasiveness underscores the importance of efficiently querying and interpreting this form of structured information. However, the intricacies of multi-dimensional queries and the vastness of data they cover often call for considerable human intervention, primarily in the construction and refinement of SQL statements \cite{1}. Consequently, the data management landscape is fraught with challenges that render the interpretation and utilization of such data a daunting task\cite{2}.

The rapidly advancing field of artificial intelligence offers potential solutions to these challenges. More specifically, recent advances in large language models present an intriguing opportunity to apply their understanding and generation capabilities to this field of research. These models, trained on diverse and massive text corpora, have showcased the capability to generate human-like text and understand the context and intricacies of the information contained within the text\cite{3}.

In this context, the current research provides an innovative take on the application of LLMs for decoding and querying tabular data. Our work aims to mitigate the need for human intervention by employing LLMs to decipher table structures, comprehend the complexities of the presented problems, and subsequently, formulate SQL queries. In contrast to traditional methodologies, this approach provides a scalable, efficient, and robust solution for handling large-scale, multi-dimensional data.

Our strategy expands on the iterative nature of LLMs, deploying them not just to generate SQL queries but also to learn from their errors and improve in an evolutionary fashion. This adaptation facilitates the handling of queries of varying complexity levels, effectively expanding the range of manageable data queries. Consequently, this approach greatly enhances the efficiency and accessibility of data management systems, providing a path to reduce the time-consuming and error-prone process of manual SQL statement construction.Moreover, we propose an innovative deployment of LLMs, effectively redefining their role in the field of data analysis and management. The potential of this pioneering methodology is confirmed through its promising performance across various datasets, exemplified by its outstanding performance on the benchmark Spider dataset.

This paper is divided into several sections detailing our research. We begin by reviewing the related work in the field, underscoring the gap that our research seeks to fill. This is followed by a detailed explanation of our methodology and how LLMs can be employed to handle tabular data. Subsequent sections detail our experimental setup, followed by an in-depth analysis of the results. Finally, we conclude with a discussion of the implications of our research, its limitations, and potential avenues for future work.

\section{Related Work}
The scope of the current research spans and contributes to several seminal works and key areas in the field, namely Large Language Models and SQL Query Generation from Natural Language, which not only serve as the foundational bedrock of our study but also constitute the backdrop against which our innovations are delineated.
\subsection{Large Language Models}

Our research is significantly founded on advancements in large language models. Specifically, the groundbreaking research on GPT-4 by OpenAI et al.\cite{4} , a language model with trillions of autoregressive parameters, has set a precedent for the capabilities of such models. They demonstrated that scaling up language models significantly improves performance on a variety of tasks, including translation, question-answering, and cloze tasks, among others. Our work extends the capabilities of these large language models, specifically LLaMA-2\cite{5}, towards understanding and generating SQL queries from table structures and problem statements, pushing the boundary of what these models can achieve.

\subsection{SQL Query Generation from Natural Language}
SQL query generation has been a focal point for many researchers\cite{6}. A notable contribution was made by Yu et al \cite{7}.with the Spider dataset and the Text-to-SQL model, significantly advancing handling of complex, cross-domain SQL queries. Yet, handling the most complex queries remained a challenge\cite{8}.

To address this, Pourreza and Rafiei\cite{9} introduced Din-SQL, applying in-context learning of Text-to-SQL through a decomposed approach with self-correction. Concurrently, Li et al.\cite{10} proposed GraphiX-T5, combining pre-trained Transformers and graph-aware layers for Text-to-SQL parsing, capitalizing on the strengths of both methodologies.Additionally, at the AAAI Conference on Artificial Intelligence in 2023, Li, Zhang, Li, and others\cite{11} unveiled ResdSQL, which decoupled schema linking and skeleton parsing for Text-to-SQL, demonstrating a new approach to the task.While these works have made substantial progress, complex query handling remains challenging. Our work uses a large language model with vast pretraining to handle intricate queries, promoting generalization across different domains.

\section{Methodology}
Our approach utilizes LLaMA-2\cite{5} as the LLM, serializing table structures and questions as inputs to the model, which then produces SQL statements for querying on the tables. Throughout this process, we employ an iterative optimization procedure and fine-tuning techniques to ensure the model's accuracy and efficiency.

\subsection{Input Construction}

\begin{figure}
\centering
\includegraphics[width=0.85\linewidth]{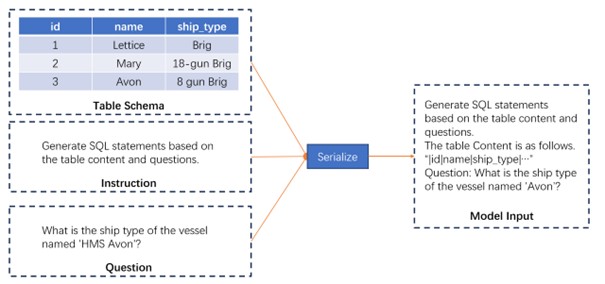}
\caption{\label{fig:1}Input Construction}
\end{figure}

Firstly, we devise a mechanism for input construction that comprises both the problem statement and the schema of the table. By concatenating these elements, we generate a robust instructional input for the model. This concatenation is not simply an arbitrary combination, but rather a well-curated blending of both the data (table schema) and the desired output (problem statement). The reasoning behind this is twofold: the schema offers the model the requisite understanding of the data structure, while the problem statement provides the goal that the query should aim to achieve. The construction process of the entire input is shown in \ref{fig:1}.
\subsection{SQL Generation}
\begin{figure}
\centering
\includegraphics[width=0.85\linewidth]{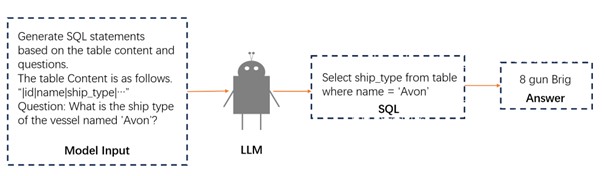}
\caption{\label{fig:2}SQL Generation}
\end{figure}

After the input is constructed, it is used to prompt the LLaMA-2 model \cite{5}. Leveraging its massive 70-billion parameter structure, the model is then tasked with generating a SQL statement as its output. This is not just a simple recitation of the input but an active interpretation of the input’s complexity and its conversion into an appropriate SQL query. The aim is for the model to take in a problem and schema as input and generate a SQL statement that accurately and efficiently answers the problem, all the while adhering to the structure of the data as represented by the schema. The generation process of the entire input is shown in \ref{fig:2}.
\subsection{Iterative Refinement}

\begin{figure}
\centering
\includegraphics[width=0.85\linewidth]{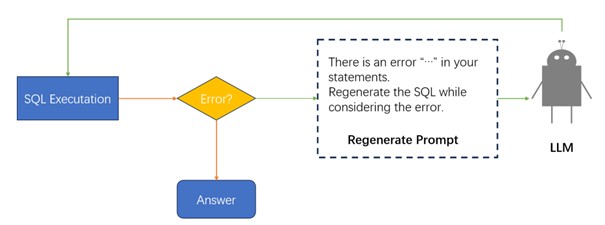}
\caption{\label{fig:3}SQL Generation}
\end{figure}

This input-output relationship forms the core of our methodology. However, to bolster the accuracy and efficiency of the SQL statements, we further incorporate an iterative refinement process. This involves the generated SQL statement being evaluated for its accuracy, with the model using this evaluation to improve its future outputs. This iterative process is not simply a loop, but rather a learning cycle where the model uses past experiences to improve future performances, effectively embodying an artificial form of evolution. The iterative refinement process of the entire input is shown in \ref{fig:3}.

\section{Experiments}
\subsection{Experimental Setup}
In order to evaluate our methodology, we conducted an empirical analysis using the Spider benchmark dataset, which provides a rich and diverse ground for testing and comparison. The experiment was run on a robust hardware configuration featuring four A100 80GB GPUs, to effectively utilize the computational power offered by the LLaMA-2 70b version.
\subsection{Performance Metrics}
We compare our work with BRIDGE\cite{11} and RESDSQL\cite{12}, with the results shown in \ref{tab:1}. Our model's performance was evaluated on two primary metrics used in the Spider dataset: Execution Accuracy and Exact Set Match\cite{13}. The results demonstrate that our methodology exhibits robust performance with an Execution Accuracy of 70.5, while achieving an Exact Set Match score of 59.3. Our method significantly outperforms the method proposed by BRIDGE, but lags slightly behind RESDSQL.

\begin{table}
\centering
\begin{tabular}{c|c|c}
Methods & Executation  Accuracy & Exact Match Score\\\hline
BRIDGE & 59.9 & 65.2 \\
RESDQL & \textbf{79.9} & \textbf{72.0} \\
\textbf{Ours} & 70.5 & 59.3
\end{tabular}
\caption{\label{tab:1}Performance Metrics}
\end{table}

\subsection{Comparative Analysis}
We divided the Spider dataset into three categories based on difficulty level: easy, medium, and hard, with data ratios of 3:2:1. As shown in \ref{tab:2}, we evaluated the EX and EM scores of RESDSQL, BRIDGE, and our proposed method on these three types of problems, where the data format is 'EX/EM'. By analyzing the results from \ref{tab:2}, we find that our method performs very well on easy problems, but does not surpass the current best solution when dealing with harder problems.

\begin{table}
\centering
\begin{tabular}{c|c|c|c}
Methods & Easy & Medium & Hard\\\hline
BRIDGE & 72.4/75.7 & 61.3/64.1 & 42.9/43.5  \\
RESDQL & 84.8/\textbf{87.0} & \textbf{81.2/83.5} & \textbf{68.1/63.6} \\
\textbf{Ours} & \textbf{85.2}/76.9 & 74.0/48.9 & 74.0/48.9
\end{tabular}
\caption{\label{tab:1}Performance Metrics}
\end{table}

\section{Conclusion}
In summary, our study highlights the flexibility of large language models, emphasizing their ability to explore various solution paths, leveraging their extensive parameter space. Despite sometimes leading to lower Exact Set Match scores, these models maintain high Execution Accuracy, making them particularly valuable in real-world scenarios where deriving the correct answer is more critical than following a predefined path\cite{14}. Our research underscores the significant capacity of large models for executing SQL queries accurately and their impressive generalizability across datasets. This underlines their potential for transformative roles in data management tasks. Looking forward, the future research direction is to enhance the ability of Large Language Models\cite{15} to understand tabular data and increase the accuracy of generating SQL queries. Our findings mark a significant stride toward leveraging the full potential of large language models.
\bibliographystyle{alpha}
\bibliography{sample}

\newcommand{\etalchar}[1]{$^{#1}$}
\begin{thebibliography}{NKP{\etalchar{+}}18}

\bibitem[DZG{\etalchar{+}}23]{14}
Xuemei Dong, Chao Zhang, Yuhang Ge, Yuren Mao, Yunjun Gao, Jinshu Lin, Dongfang
  Lou, et~al.
\newblock C3: Zero-shot text-to-sql with chatgpt.
\newblock {\em arXiv preprint arXiv:2307.07306}, 2023.

\bibitem[KSHL20]{8}
Hyeonji Kim, Byeong-Hoon So, Wook-Shin Han, and Hongrae Lee.
\newblock Natural language to sql: Where are we today?
\newblock {\em Proceedings of the VLDB Endowment}, 13(10):1737--1750, 2020.

\bibitem[LSX20]{12}
Xi~Victoria Lin, Richard Socher, and Caiming Xiong.
\newblock Bridging textual and tabular data for cross-domain text-to-sql
  semantic parsing.
\newblock {\em arXiv preprint arXiv:2012.12627}, 2020.

\bibitem[LZLC23]{11}
Haoyang Li, Jing Zhang, Cuiping Li, and Hong Chen.
\newblock Resdsql: Decoupling schema linking and skeleton parsing for
  text-to-sql.
\newblock In {\em Proceedings of the AAAI Conference on Artificial
  Intelligence}, volume~37, pages 13067--13075, 2023.

\bibitem[NKP{\etalchar{+}}18]{2}
Fatemeh Nargesian, Udayan Khurana, Tejaswini Pedapati, Horst Samulowitz, and
  Deepak Turaga.
\newblock Dataset evolver: An interactive feature engineering notebook.
\newblock In {\em Proceedings of the AAAI Conference on Artificial
  Intelligence}, volume~32, 2018.

\bibitem[Ope23]{4}
OpenAI.
\newblock Gpt-4 technical report, 2023.

\bibitem[PR23a]{9}
Mohammadreza Pourreza and Davood Rafiei.
\newblock Din-sql: Decomposed in-context learning of text-to-sql with
  self-correction.
\newblock {\em arXiv preprint arXiv:2304.11015}, 2023.

\bibitem[PR23b]{10}
Mohammadreza Pourreza and Davood Rafiei.
\newblock Din-sql: Decomposed in-context learning of text-to-sql with
  self-correction.
\newblock {\em arXiv preprint arXiv:2304.11015}, 2023.

\bibitem[RBE{\etalchar{+}}20]{1}
Alexander Ratner, Stephen~H Bach, Henry Ehrenberg, Jason Fries, Sen Wu, and
  Christopher R{\'e}.
\newblock Snorkel: Rapid training data creation with weak supervision.
\newblock {\em The VLDB Journal}, 29(2-3):709--730, 2020.

\bibitem[RSR{\etalchar{+}}20]{3}
Colin Raffel, Noam Shazeer, Adam Roberts, Katherine Lee, Sharan Narang, Michael
  Matena, Yanqi Zhou, Wei Li, and Peter~J Liu.
\newblock Exploring the limits of transfer learning with a unified text-to-text
  transformer.
\newblock {\em The Journal of Machine Learning Research}, 21(1):5485--5551,
  2020.

\bibitem[RSR{\etalchar{+}}22]{6}
TJ~Revanth, K~Venkat Sai, R~Ramya, Renusree Chava, V~Sushma, and BS~Ramya.
\newblock Nl2sql: Natural language to sql query translator.
\newblock In {\em Emerging Research in Computing, Information, Communication
  and Applications: ERCICA 2020, Volume 2}, pages 267--278. Springer, 2022.

\bibitem[TMS{\etalchar{+}}23]{5}
Hugo Touvron, Louis Martin, Kevin Stone, Peter Albert, Amjad Almahairi, Yasmine
  Babaei, Nikolay Bashlykov, Soumya Batra, Prajjwal Bhargava, Shruti Bhosale,
  et~al.
\newblock Llama 2: Open foundation and fine-tuned chat models.
\newblock {\em arXiv preprint arXiv:2307.09288}, 2023.

\bibitem[YZY{\etalchar{+}}18]{7}
Tao Yu, Rui Zhang, Kai Yang, Michihiro Yasunaga, Dongxu Wang, Zifan Li, James
  Ma, Irene Li, Qingning Yao, Shanelle Roman, et~al.
\newblock Spider: A large-scale human-labeled dataset for complex and
  cross-domain semantic parsing and text-to-sql task.
\newblock {\em arXiv preprint arXiv:1809.08887}, 2018.

\bibitem[ZYK20]{13}
Ruiqi Zhong, Tao Yu, and Dan Klein.
\newblock Semantic evaluation for text-to-sql with distilled test suites.
\newblock {\em arXiv preprint arXiv:2010.02840}, 2020.

\bibitem[ZZL{\etalchar{+}}23]{15}
Wayne~Xin Zhao, Kun Zhou, Junyi Li, Tianyi Tang, Xiaolei Wang, Yupeng Hou,
  Yingqian Min, Beichen Zhang, Junjie Zhang, Zican Dong, et~al.
\newblock A survey of large language models.
\newblock {\em arXiv preprint arXiv:2303.18223}, 2023.

\end{thebibliography}

\end{document}